\documentclass{article}
\usepackage{graphicx} 
\usepackage[a4paper, total={6in, 8in}]{geometry}

\usepackage{amsmath}
\usepackage{amsfonts}
\usepackage{amssymb}
\usepackage{url}

\usepackage{subfigure}
\usepackage[toc,page]{appendix}

\usepackage{natbib}
\usepackage{authblk}

\DeclareMathOperator{\E}{\mathbb{E}}

\usepackage{algorithm}
\usepackage{algpseudocode}

\title{Decorrelated Soft Actor-Critic\\for Efficient Deep Reinforcement Learning}
\author{Burcu Küçükoğlu, Sander Dalm, Marcel van Gerven}
\affil{Department of Machine Learning and Neural Computing, Donders Institute for Brain, Cognition and Behaviour, Radboud University, 
The Netherlands}
\date{}

\begin{document}

\maketitle

\begin{abstract}

The effectiveness of credit assignment in reinforcement learning (RL) when dealing with high-dimensional data is influenced by the success of representation learning via deep neural networks, and has implications for the sample efficiency of deep RL algorithms. Input decorrelation has been previously introduced as a method to speed up optimization in neural networks, and has proven impactful in both efficient deep learning and as a method for effective representation learning for deep RL algorithms. We propose a novel approach to online decorrelation in deep RL based on the decorrelated backpropagation algorithm that seamlessly integrates the decorrelation process into the RL training pipeline. Decorrelation matrices are added to each layer, which are updated using a separate decorrelation learning rule that minimizes the total decorrelation loss across all layers, in parallel to minimizing the usual RL loss. We used our approach in combination with the soft actor-critic (SAC) method, which we refer to as decorrelated soft actor-critic (DSAC). Experiments on the Atari 100k benchmark with DSAC shows, compared to the regular SAC baseline, faster training in five out of the seven games tested and improved reward performance in two games with around 50\% reduction in wall-clock time, while maintaining performance levels on the other games. These results demonstrate the positive impact of network-wide decorrelation in deep RL for speeding up its sample efficiency through more effective credit assignment.

\end{abstract}

\section{Introduction}
Reinforcement learning (RL) is a framework for learning effective control policies in sequential decision making tasks, based on feedback from the task environment~\citep{Sutton1998rlintro}. The feedback is received in the form of rewards, following the actions taken by an artificial agent at every time step. The key in learning an optimal action sequence lies in correctly identifying the impact of an action on future rewards. Such credit assignment is especially challenging when dealing with complex high dimensional data like images, which make the learning of the state representations critical for RL's success. The use of deep neural networks as function approximators facilitated RL's achievement in representation learning~\citep{minh2013playingataridrl, mnih2015human}. However, RL needs still a lot of interactions to be able to solve a task, which makes it notoriously sample inefficient.

Targeting the issue of sample efficiency has long been a focus of RL research~\citep{kaiser2020simple, haarnoja2018icml, sac2018haarnoja, küçükoğlu2024efficient}. A common characteristic of all these RL approaches is their use of stochastic gradient descent (SGD) through the backpropagation (BP) algorithm~\citep{linnainmaa1976bp, werbos1974bp} for training of the deep neural network (DNN) structures.
When implementing SGD through BP naively, convergence can be slowed down when correlations are present in a layer's inputs~\citep{lecun200efficientbp}. A method to remedy this inefficiency is to decorrelate each network layer's input before applying the forward weights. For an in-depth explanation of this, see~\citet{ahmad2024correlations}.

Decorrelation has been shown to speed up training for both limited-depth fully-connected networks, demonstrated by 78\% epoch wise training time reduction on CIFAR-10~\citep{ahmad2023constrained}, and for large-scale DNNs with convolutional layers and residual connections via the novel decorrelated backpropagation (DBP) algorithm, reducing training wall-clock time of ResNet-18 on ImageNet by 59\%~\citep{dalm2024efficientdeeplearningdecorrelated}.
Building on these works, we propose to transfer the benefits of decorrelation to the RL setting to improve sample efficiency. We hypothesize that deep RL with decorrelation across its neural network layers' inputs would yield a speedup in training and improved performance via more 
efficient gradient descent. This would in turn allow more effective representation learning 
and faster convergence to effective policies. 
This increase in training efficiency may also help reduce carbon emissions generated by RL training of large networks.

To this end, we propose to combine decorrelated backpropagation with the state-of-the-art soft actor-critic (SAC) algorithm~\citep{sac2018haarnoja} to achieve  more sample efficient RL. We refer to this approach as decorrelated soft actor-critic (DSAC).
The decorrelation process is based on the DBP algorithm~\citep{dalm2024efficientdeeplearningdecorrelated} with minor modifications for the RL setting. Every fully-connected and convolutional layer input of the 
DNN to be decorrelated is
preceded by a linear transformation that applies decorrelation on the inputs via multiplication by a decorrelating matrix.
The decorrelating matrix is updated simultaneously with the forward weights, but using a separate decorrelation learning rule. The optimization is achieved through the minimization of the total decorrelation loss from all layers and the RL loss.

Our method differs fundamentally from previous approaches to decorrelation in RL. While early work was similarly motivated by simultaneous online learning of good representation and task control in a single phase process, decorrelation was implemented via regularization of the DQN loss based on the correlation in the last hidden layer's latent features only~\citep{mavrin2019drlwithdecorr, mavrin2019effdecorgramian}.
\citet{lee2023decorunsupreprlearn} applied decorrelation only in an offline pretraining phase for unsupervised representation learning, where an encoder predicting future states decorrelated features encoded in the latent space with a regularizing decorrelation loss next to the similarity loss. A pretrained encoder was then used to train the policy separately.~\citet{huang2023generalizabledecorsaliency} also used SAC as the base algorithm, yet with only fully-connected layers. Their decorrelation implementation acted on a latent feature vector output by an encoder separate from the SAC networks for actor and critic. Besides, additional support from saliency maps and a classifier was used to guide the decorrelation process as their approach focused only on visual RL tasks.

In contrast, our work applies 
network-wide
decorrelation in an online fashion to all the layer inputs of all layers of 
a
trained 
network
within the RL context. Given the use of the SAC algorithm with separate actor and critic architectures, decorrelation in even multiple DNNs simultaneously can be accommodated, along with the usual RL training pipeline, in a single phase process that includes representation learning without requiring pretraining. These DNNs also include convolutional layers, going beyond mere multi-layer perceptrons. In our work, decorrelation acts on every layer input
of the actor network,
rather than on some latent representation learned by a whole encoder, or only on the last hidden layer as a representative for correlation computation. As decorrelation is directly applied to network layer inputs, no additional architecture like a separate encoder is needed. Furthermore, no complex additions that require extra computation of saliency maps or classifier training are needed, since our decorrelation approach is not constrained to visual RL tasks.
Finally, instead of 
minimizing correlations by regularizing the network's loss function, we update the decorrelation matrices in an entirely separate, though simultaneous, process using its own learning rule.  

We thus contribute an iterative, online decorrelation method that is fully integrated with the network training and speeds up convergence of RL algorithms through more efficient representation learning.

\section{Methods}
\subsection{Decorrelated Backpropagation}
\label{sec:decbp}
Decorrelated backpropagation (DBP)~\citep{dalm2024efficientdeeplearningdecorrelated} is an alternative algorithm to standard backpropagation for efficient training of neural networks. It utilizes an iterative decorrelation procedure~\citep{dalm2024efficientdeeplearningdecorrelated, ahmad2023constrained} for decorrelation of layer inputs in the network. This is achieved by introducing an additional matrix to each layer which decorrelates inputs before they are multiplied by the forward weights. Decorrelation of each layer input is implemented independently as a linear transformation via
\begin{equation}
\mathbf{x = Rz} 
\end{equation}
where $\mathbf{z}$ is the raw layer input that is multiplied by the decorrelating matrix $\mathbf{R}$ to get the decorrelated input $\mathbf{x}$. The decorrelated input is then passed on to the original network layer to accomplish the forward pass, which gives the output of that layer as
\begin{equation}
\mathbf{y} = f\mathbf{(Wx)}= f\mathbf{(WRz)} = f\mathbf{(Az)}
\end{equation}
where output $\mathbf{y}_{l}$ of a layer serves as the correlated input $\mathbf{z}_{l+1}$ of the next layer. Note that $f$ is a non-linear transformation and $\mathbf{Wx}$ 
can be any multiplication by forward weights (e.g. fully-connected, convolutional, recurrent, etc).

The decorrelating matrix $\mathbf{R}$ for each layer is initialized as the identity matrix and then trained independently 
by applying:
\begin{equation}
\mathbf{R} \leftarrow \mathbf{R} - \eta \mathbf{C} \mathbf{R}
\end{equation}
where $\eta$ is a learning rate and  $\mathbf{C}$ is the empirical off-diagonal correlation of the decorrelated input 
with itself, given by
\begin{equation}
\mathbf{C} =  \E \left[ \mathbf{x} \mathbf{x}^\top \right]  -\text{diag}\left( \E \left[ \mathbf{x}^2\right]  \right) 
\end{equation}
where $\text{diag}\left(\E \left[ \mathbf{x}^2\right]  \right)$ is the diagonal of the expectation of the correlation matrix $\mathbf{x} \mathbf{x}^\top$.
We are thus aiming to minimize the off-diagonal elements of $\mathbf{x} \mathbf{x}^\top$.
The above learning rule minimizes a decorrelation loss $d$ for layer $l$ given by
\begin{equation}
d_l = \sum_{i=1}^n\sum_{j=1}^n c_{ij}^2 
\end{equation}
where ${c}_{ij}$ are the elements of $\mathbf{C}_l$, 
the correlation matrix of layer $l$. Thus, the total decorrelation loss $D$ for the network is
\begin{equation}    
D = \sum_{l=1}^{L} d_l
\end{equation}
where $L$ is the total number of layers.

Note that while the above formulas give a general description of our decorrelation procedure, there are implementation differences between fully-connected and convolutional layers. For convolutional layers, slight modifications are required to minimize the extra computational overhead that is introduced with the addition of decorrelating matrices for each network layer.

\subsection{Reducing the Computational Burden}
In order to decrease the computational overhead for the decorrelation process in the convolutional layers, three modifications are used, following~\citet{dalm2024efficientdeeplearningdecorrelated}. The first measure addresses the high computational cost of decorrelating large feature maps by applying decorrelation on local image patches instead of the whole input. This enables the decorrelated input to be transformed via $1 \times 1$ convolutional kernels for the forward pass.

The second modification replaces the expensive multiplication $\mathbf{W(Rz)}$ with the cheaper multiplication $\mathbf{(WR)z}$, via an initial condensation of matrices $\mathbf{W}$ and $\mathbf{R}$ into matrix $\mathbf{A = WR}$, to be only later multiplied by the raw input $\mathbf{z}$ for the forward pass. This modification capitalizes on the lower dimensionality of $\mathbf{W}$ compared to that of $\mathbf{z}$, as $\mathbf{W}$ does not have a batch dimension~\citep{dalm2024efficientdeeplearningdecorrelated}.

The third modification reduces the cost of computing the correlation matrix during the update calculation by basing it on a random sample of the input.
~\citet{dalm2024efficientdeeplearningdecorrelated} found that sampling 10\% of the input samples is sufficient to learn the data's correlational
structure with minimal loss of accuracy. To this end, in this study, an alternative implementation of downsampling was used, to ensure sufficient samples for performance in our use case. 
A subset of input samples were randomly chosen to compute the decorrelation update. The sample size $n$ was computed as
\begin{equation}
n = \max\left(10,b D_r p^{-1} + 1 \right)
\end{equation}
rounded to the nearest integer, where $D_r$ is the total dimensionality per input patch, $p$ the number of image patches and $b$ a scaling factor determining the degree of downsampling. The justification for this sampling method is that downsampling should be less aggressive if the dimensionality of $R$ is high relative to the number of patches, while always keeping a minimum of 10 samples.

\subsection{Integrating Decorrelation into Soft Actor-Critic}
Soft Actor-Critic (SAC)~\citep{sac2018haarnoja} is a state-of-the art deep RL algorithm known for its stable convergence properties and sample efficiency. It is differentiated by the combined use of three ingredients: the separate architectures for policy and value function networks as an actor-critic method, the off-policy gradient updates via the reuse of data from a replay buffer, and the addition of maximum entropy learning in its objective function. 

The SAC agent is composed of one network approximating the policy function 
and two networks approximating two soft Q-functions. 
Here,  
policy iteration is formulated via the evaluation of the current policy's Q-function and the policy's off-policy gradient update.

The soft Q-functions are trained independently with help from two target soft Q-functions.
The policy is updated via minimization of the KL-divergence between the policy distribution and the minimum of the soft Q-functions.
Alongside the expected return, the entropy is maximized to encourage exploration and stability by establishing a minimum constraint on the average entropy of the policy and through the gradient-based tuning of a temperature parameter $\alpha$ weighing the entropy against the reward maximization objective. 

For the purposes of our research, a version of SAC for discrete action spaces was used, following the approach of~\citet{christodoulou2019sacdiscrete}. 
Discrete SAC differs from the original algorithm~\citep{sac2018haarnoja} that has been designed for continuous action spaces.

\subsubsection{Soft Q-networks}
In discrete SAC, Q-networks receive the state as the only input, and output Q-values for each action separately. 
Each network is composed of 3 convolutional layers, whose consecutive output is flattened to be followed by 2 fully-connected layers. After each of these 5 layers, a Leaky ReLU activation function is applied, following after the previously described decorrelating transforms~\citep{ahmad2023constrained, dalm2024efficientdeeplearningdecorrelated}. This is different than the ReLU nonlinearity used in the original SAC implementation~\citep{sac2018haarnoja}, due to performance improvement observed for the decorrelation procedure with this change in the RL setting. 

The three convolutional layers use kernel sizes of 8, 4, 3 and strides 4, 2, 1 respectively with zero padding, while having output channels of 32, 64, 64 respectively. As for fully-connected layers, the output channels are respectively 512 and the number of actions in order to output a Q-value for each action available in the task.

\subsubsection{Policy Network}
Due to the discrete action space setting, the policy is modeled as a categorical distribution with the policy network outputting the unnormalized log probabilities of actions available that parameterize the action distribution. The action taken is determined by sampling from this distribution based on the action probabilities. Normalized log probabilities are also produced after a softmax application to network output, to be used in policy update and Q-function computations. 

The network has the same architecture as a Soft Q-network, yet has its own separate architecture and initialization with its own parameters.
Note that with the incorporation of the decorrelation process, all these layers are modified to have an additional multiplication with the decorrelating matrix for the decorrelation transformation before their own forward pass, hence after the application of an activation function in case of a previous layer existing. These decorrelation layers are differently implemented depending on whether they decorrelate inputs for a convolutional or fully-connected layer, as mentioned in Section~\ref{sec:decbp}.

\subsection{Optimization of Discrete SAC via Decorrelated BP}
As an off-policy algorithm, SAC starts by sampling random actions in the environment to first fill the replay buffer with sufficient state transition data to be reused in the weight updates. When the replay buffer has enough samples, learning starts. Each iteration of the algorithm involves an environment step, followed by a gradient step. The environment step is taken with the action sampled from the policy network's output distribution. At every gradient step, a batch of transition data
is sampled from the replay buffer to be used for the calculation of loss terms involved in the SAC algorithm. 
After the separate updates of two soft Q-functions, the policy function, and the temperature parameter $\alpha$, finally the decorrelating matrices that serve as the weights for the decorrelation transformation are updated for each layer of the decorrelated policy network. The pseudocode is provided in Algorithm \ref{alg:dbp_sac_discrete}.

Each soft Q-function is updated based on the mean squared error (MSE) loss between its Q-function output and the commonly used target Q-values for both networks, which uses the minimum of the two soft Q-targets in the calculation of the next Q-values as follows:
\begin{equation}
J_Q(\theta) = \E_{(s_t,a_t) \sim \mathcal{D}}\left[\frac{1}{2}\biggl(Q_{\theta}(s_t) - \Bigl(r(s_t,a_t) + \gamma \pi(a'|s')^T \bigl(\smash{\displaystyle\min_ {i=1,2}} Q_{\bar{\theta}_i}(s') - \alpha \log \pi(\cdot |s')\bigr)\Bigr)\biggr)^2\right]
\end{equation}
where $\mathcal{D}$ is the replay buffer, $r$ rewards, $\pi$ policy, $s'$  next states, $a'$ next actions, and $\gamma$ the discount factor on the next Q-values. Parameters $\bar{\theta}$ of a target Q-function are updated 
based on the exponential moving average of the parameters $\theta$ of the respective soft Q-function.

The policy network is updated based on the KL-divergence between the policy distribution and the minimum of the soft Q-functions, scaled by the action probabilities, thus giving the policy loss:

\begin{equation}
J_\pi(\phi) = \E_{s_t \sim \mathcal{D}}\left[ \pi(a_t|s_t)^T\Bigl(\alpha \log(\pi(a_t|s_t)) -\smash{\displaystyle\min_ {i=1,2}} Q_{\theta_i}(s)\Bigr)\right]
\end{equation}

The entropy adjusting temperature loss, whose optimizer is parametrized by the log of the temperature parameter $\alpha$, is given by  

\begin{equation}
J_\alpha =  \pi(a_t|s_t)^T \left[-\alpha \Bigl( \log(\pi(s_t)) + \mathcal{H} \Bigr)\right]
\end{equation}
where $\mathcal{H}$ is the desired minimum entropy target. Notice that in all loss formulas a scaling with the action probabilities is applied due to the discrete action setting, in order to reduce the variance of the gradient.

Finally, the total decorrelation loss $D_{total}$ in the SAC agent can be calculated by an addition of the decorrelation losses of all networks that make use of the decorrelation procedure, which corresponds to total decorrelation loss in the policy network in our implementation, with
\begin{equation}
D_{total} =  \sum_{k=1}^N 
D_{k}  
\label{eq:totaldecorloss}
\end{equation}
where $D_{k}$ is the sum of decorrelation losses across the respective network layers, and $N$ number of networks decorrelated. We refer to the resulting approach as decorrelated soft actor-critic (DSAC).

\begin{algorithm}
\caption{Decorrelated Soft Actor-Critic Algorithm}
\label{alg:dbp_sac_discrete}
\begin{small}
\begin{algorithmic}
    \State Initialize $Q_{\theta_1} : S \rightarrow \mathbb{R}^{|A|}, 
Q_{\theta_2} : S \rightarrow \mathbb{R}^{|A|},
\pi_{\phi} : S \rightarrow [0, 1]^{|A|}$
\Comment{Initialize networks}
    \State $\mathbf{R}_l \gets \mathbf{I}$ \Comment{Initialize decorrelating matrices as identity matrix for each layer of policy $\pi_\phi$}
    \State $\bar{\theta}_1 \gets \theta_1, \bar{\theta}_2 \gets \theta_2$ \Comment{Initialize target network weights}
    \State $\mathcal{D} \gets \emptyset$ \Comment{Initialize an empty replay buffer}

    \For{each iteration}
        \For{each environment step}
            \State $a_t \sim \pi_\phi(a_t | s_t)$ \Comment{Sample action from the policy}
            \State $s_{t+1} \sim p(s_{t+1} | s_t, a_t)$ \Comment{Sample transition from the environment}
            \State $\mathcal{D} \gets \mathcal{D} \cup \{(s_t, a_t, r(s_t, a_t), s_{t+1})\}$ \Comment{Store the transition in the replay pool}
        \EndFor

        \For{each gradient step}
            \State $\theta_i \gets \theta_i - \lambda_Q \nabla_{\theta_i} \hat{J}_Q(\theta_i)$ for $i \in \{1, 2\}$ \Comment{Update the Q-function parameters}
            \State $\phi \gets \phi - \lambda_\pi \nabla_{\phi} \hat{J}_\pi(\phi)$ \Comment{Update policy weights}
            \State $\alpha \gets \alpha - \lambda \nabla_{\alpha} \hat{J}(\alpha)$ \Comment{Adjust temperature}
            \State $\bar{\theta}_i \gets \tau \theta_i + (1 - \tau) \bar{\theta}_i$ for $i \in \{1, 2\}$ \Comment{Update target network weights}
            \State $\mathbf{R}_l \leftarrow \mathbf{R}_l - \eta \mathbf{C}_l \mathbf{R}_l$ \Comment{Update decorrelating matrices for each layer of $\pi_\phi$}
        \EndFor
    \EndFor

    \State \textbf{Output:} $\theta_1, \theta_2, \phi, \mathbf{R}_l$ \Comment{Optimized parameters}
\end{algorithmic}
\end{small}
\end{algorithm}

\subsection{Empirical validation of DSAC}
For empirical validation of DSAC, we utilize the Atari 100k benchmark that is commonly used when testing sample-efficiency of RL algorithms. We follow the initial proposal of~\citet{kaiser2020simple}, and discrete SAC evaluations by~\citet{christodoulou2019sacdiscrete}, where agents are trained for 100k environment steps, corresponding to 400k frames with frame skipping of 4. We focus on seven environments with higher difficulty levels and 18 possible actions to get the full action space, also given existing performance scores for some games on the previous discrete SAC implementation~\citep{christodoulou2019sacdiscrete}. 
The games of focus are PrivateEye, Frostbite, Seaquest, Alien, BankHeist, ChopperCommand and BattleZone, from the most to least difficult, based on DQN's performance in these games relative to human performance~\citep{mnih2015human}. 
The original Atari RGB color frames of 210 by 160 pixels were transformed into the commonly used 84 by 84 pixels in grayscale. With the application of typical four-frame stacking~\citep{machado18}, the final input had four channels of $84 \times 84$. 
Sticky actions were used with an action repeat probability of 0.25 for injection of stochasticity into the environment as advised by~\citet{machado18}. No limit on episode duration was imposed, based on the suggestion of~\citet{toromanoff2019deep}. 

We compare the performance of DSAC against a typical BP-based discrete SAC algorithm. 
For both algorithms, a grid search for hyperparameters was performed, where the learning rate for the SAC algorithm and the batch size for sampling from the replay buffer were treated as hyperparameters, due to their observed impact on the performance. For DSAC, additionally the learning rate for the decorrelation update of the policy network was treated as a hyperparameter. 
Identical learning rates were used for the SAC components, namely the two soft Q-networks, the policy network and for the updates of the entropy adjusting temperature parameter. The learning rates experimented with were \{$3 \times 10^{-5}$, $1 \times 10^{-4}$, $3 \times 10^{-4}$\} and \{$1 \times 10^{-4}$, $1 \times 10^{-3}$, $1 \times 10^{-2}$\} for the SAC components and the decorrelation of the policy network respectively. The batch sizes experimented with were \{64, 256\}, after the choice of these values by the original discrete SAC~\citep{christodoulou2019sacdiscrete} and SAC~\citep{sac2018haarnoja} papers respectively. The grid search results on final reward performance can be seen in Appendix \ref{sec.gridsearch}.
For the scaling coefficient $b$ used in downsampling of convolutional DBP layers, a value of 9 was observed to be most effective based on preliminary analysis, in terms of maintaining task performance while still decreasing computational costs. Therefore this value was kept fixed across experiments.

The algorithms' results were compared at the end of 100k training steps, thus a total of 120000 steps given initial random steps to fill the replay buffer. All hyperparameters used in this study for DBP and the discrete SAC algorithm can be found listed in Appendix \ref{sec.hyperparams}.
For additional implementation details, please refer to Appendix \ref{sec.hardware}. The code required to reproduce all simulations are available at \url{https://github.com/burcukoglu/Decorrelated_SAC.git}.

\section{Results}

\subsection{Training Time and Reward Comparison}

\begin{figure*}[!ht]
\centering
\includegraphics[width=\textwidth]{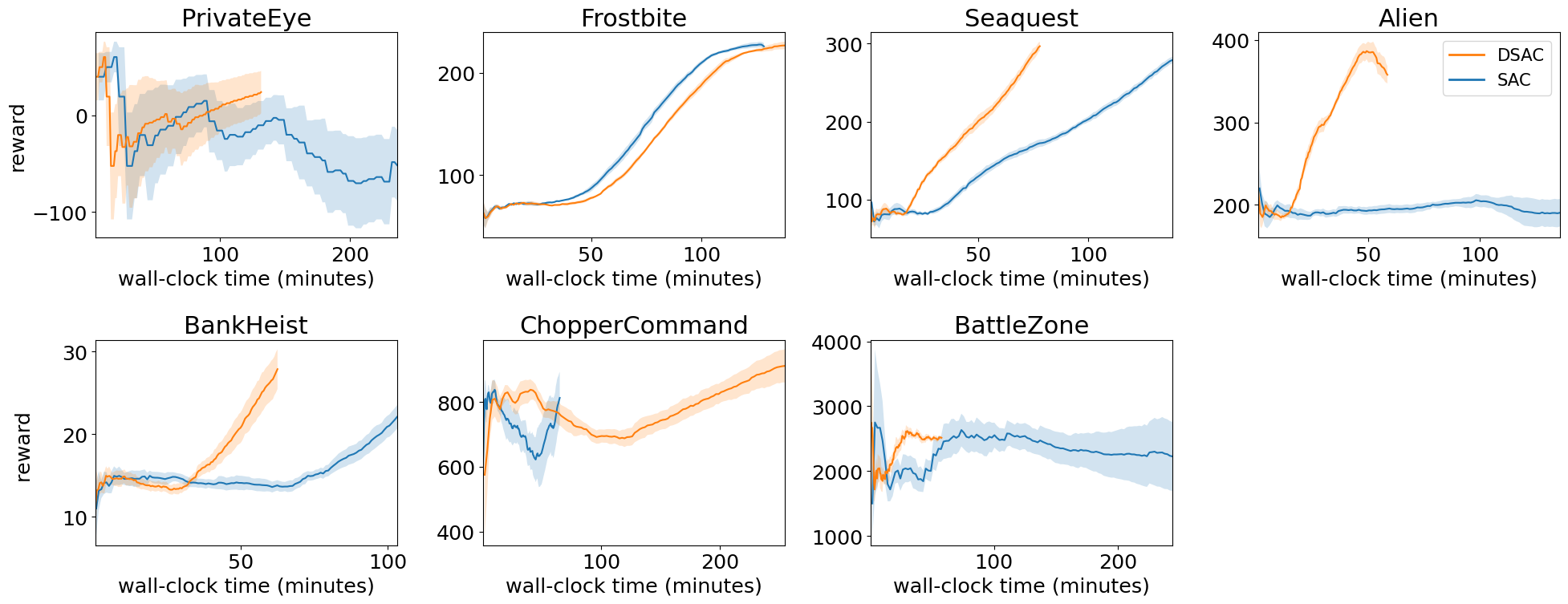}
\caption{DSAC speeds up training w.r.t. SAC in 
5
of the 7 games tested for up to 
76\%,
maintaining performance in all 7 and even improving performance in 
2
games (Seaquest and Alien). Both algorithms use the best hyperparameter configuration for each game, with results averaged over 5 seeds. The shaded area shows the standard error of the mean. The games on which DSAC is faster than SAC are the games where DSAC uses a lower batch size (64) than SAC (256). 
For ChopperCommand, DSAC is slower due to higher batch size (256) against SAC (64). For Frostbite the time difference is small as SAC and DSAC use the same batch size (256).} 
\label{fig:timeadvantage}
\end{figure*}

Figure \ref{fig:timeadvantage} compares the reward performance of DSAC against a SAC baseline across the wall-clock times during the training for a fixed number of steps. The algorithms plotted are based on the best configurations that emerged for each algorithm per each environment as a result of the hyperparameter search.
The learning rates and batch sizes for these best configurations can be seen in Appendix \ref{sec.topconfigurations}. DSAC completes training for the same number of steps in a shorter wall-clock time in 
5
out of 7
games tested, demonstrating efficiency benefits of the network-wide decorrelation. It also significantly outperforms the SAC baseline ($p<$ 0.05, $N$ = 10) in 
2 of the games tested, showing an 86\% performance gain in Alien and 6\% in Seaquest. 

Alien and Seaquest are also the games where DSAC provides a time advantage, training 44-57\% faster. 
Thus,
 large speed benefits as a result of decorrelation translate into performance benefits, potentially due to effective credit assignment. On the remaining games, DSAC and SAC baseline have similar performance levels ($p>$ 0.05, $N$ = 10). 
Please refer to Appendix \ref{sec.mainresults} for reward curves based on training steps, and their comparison to the performance of other discrete SAC benchmarks.

\subsection{Effect of Batch Size}
When we look closely at the hyperparameters for these 
plotted algorithms based on 
best configurations, we see that only in ChopperCommand  DSAC has a higher batch size than the SAC baseline, explaining the 
slowness of DSAC for 
that game. 
Frostbite is also the only case where SAC and DSAC have identical batch-sizes, which leads DSAC to have a slightly longer training time for this game due to the additional computation for decorrelation updates. 
In all the games where DSAC training is faster than SAC, DSAC has a lower batch size than SAC.
While this may indicate an impact of the lower batch size on the success of DSAC, our experimentation shows that it is not the sole factor in its success (see Appendix \ref{sec.resultsalldecorrelated}). 
With an increase in the batch size, DSAC indeed provides less time advantage compared to the lower batch size, yet still 
is able to 
reduce the wall-clock time against the SAC baseline 
when run with an identical batch size.
Similarly, despite identical batch sizes, DSAC is still able to outperform SAC.
These results show that while a lower batch size is beneficial for reducing training time, the benefits of decorrelation do not entirely depend on it.

\subsection{Learning Rate Comparison}
As for learning rates, each game had its own best configuration. The lower learning rates provided the top performers of the SAC baseline, with \{$3 \times 10^{-5}$, $1 \times 10^{-4}$\} being equally chosen, thus making it hard to converge on one choice for all experiments. These were almost always accompanied by a batch size of 256. In contrast, \{$3 \times 10^{-4}$\} was never chosen for the top configurations. For DSAC's SAC learning rate, \{$1 \times 10^{-4}$, $3 \times 10^{-4}$\} were common choices, with \{$3 \times 10^{-4}$\} always being accompanied by a batch size of 64 as the more common choice for batch size in highest ranking DSAC configurations. As for decorrelation learning rates, the choices in order of popularity were \{$1 \times 10^{-4}$, $1 \times 10^{-2}$, $1 \times 10^{-3}$\}. These show that a unique decorrelation learning rate might provide more value depending on the environment.

\subsection{Impact on Decorrelation Loss}
\begin{figure*}[!htb]
\centering
\includegraphics[width=\textwidth]{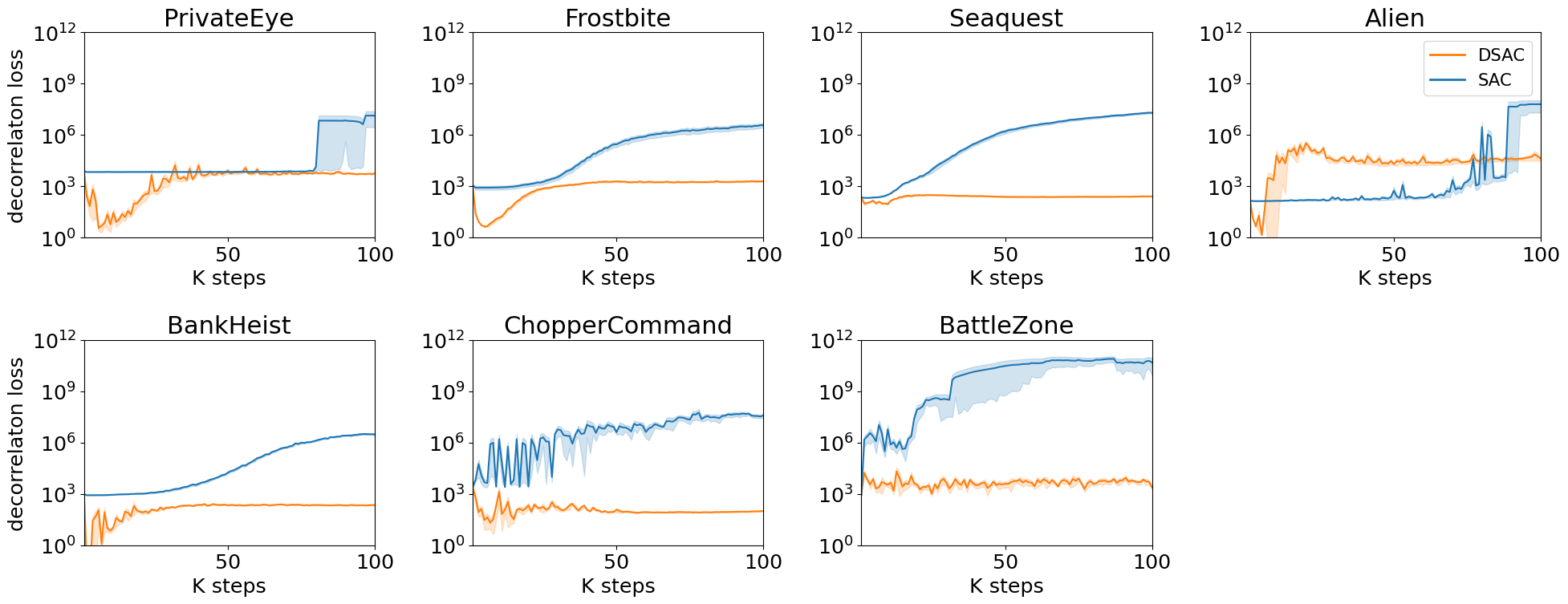}
\caption{Decorrelation loss for the policy network (shown in log scale) constitutes the main difference across the compared algorithms, which is successfully minimized throughout the learning in DSAC, despite steady increase for the SAC baseline, where it reaches values beyond $10^6$. Plots are based on a SAC baseline trained with similar architectural conditions with DSAC, which differs from the baseline in the main results, to enable a fair comparison of the decorrelation losses.}
\label{fig:policydecorloss}
\end{figure*}

To further investigate the impact of decorrelation, an analysis of 
the decorrelation losses can be made. 
Figure \ref{fig:policydecorloss} displays the 
total
decorrelation loss for the policy network, 
which was
subject to a hyperparameter search for its learning rate. 
Here the top performing configurations found in the hyperparameter search are plotted for the algorithms with the loss values in log scale. 
Among the 
curves comparing DSAC and the SAC baseline, the most obvious and systematic difference emerges in the minimization of the decorrelation loss for the policy network, which is consistently kept at a minimum for DSAC throughout training, while it increases steadily for the SAC baseline towards the final steps. These results show successful decorrelation in the DSAC policy networks.
While the policy network learns increasingly correlated features with BP, decorrelation via DBP helps mitigate this effect by ensuring minimal correlation throughout training. 

\subsection{Impact on RL Loss}
In Appendix~\ref{sec.losscurves}, other loss curves from the SAC algorithm can be observed. The increase in the SAC baseline's non-optimized decorrelation loss for the policy network seems to coincide with the convergence of the actor loss in games like BankHeist and Seaquest, signaling the exploitation stage for the algorithm. The alpha loss for DSAC on the other hand reaches convergence faster than the SAC baseline consistently. This shows that decorrelation helps in reaching the entropy maximization objective faster, by encouraging exploration via minimization of the correlation in features. Based on this, the minimal levels of the decorrelation loss can be seen as a signal of how novel the visited states are. This way, decorrelation via DBP contributes to the SAC algorithm's entropy maximization goals.

\section{Discussion}
We have proposed DSAC as a novel approach to online network-wide decorrelation in deep RL based on the DBP algorithm that applies decorrelation across its neural network layer inputs before their forward pass, for effective representation learning and improved sample efficiency in RL. An application of DBP to the discrete SAC algorithm has demonstrated faster training in wall-clock time in the majority of the Atari games tested, which converted to performance benefits in a few games against a BP-based discrete SAC baseline. 
DSAC showed successful minimization of the decorrelation loss for the policy network where decorrelation was applied, whereas the SAC baseline displayed high correlation of features. 
Our results suggest that network-wide decorrelation indeed shows advantage in effective representation learning by removing correlations, thus improving sample efficiency by speeding up the deep RL training.

Deep RL has been previously shown to learn features that are highly correlated, which cause slow learning curves~\citep{mavrin2019drlwithdecorr}.~\citet{mavrin2019effdecorgramian} observed that applying decorrelation in deep RL through regularization leads to richer representations with higher number of useful features learned. By allowing the agent to better understand the effect of changing features on its decisions, decorrelation has been proposed as an approach to achieve generalization by RL agents across environments~\citep{huang2023generalizabledecorsaliency}.

Here we built on these findings through a fully-integrated decorrelation process to the training of the RL pipeline, which differs fundamentally from previous approaches. Next to the usual RL updates, we apply separate updates to additional decorrelation matrices added to each layer of a DNN, based on a decorrelation learning rule. Thus we present an approach to simultaneous online learning of good representations and task control in a single phase process, without the need for regularization, pretraining, use of additional network architectures or a focus on a single layer's latents. Our method of decorrelation provides an advantage by being applicable to any kind of existing algorithm or DNN architecture, with the possibility for an application in any type of network layer, demonstrated here with convolutional and fully-connected layers. Furthermore, through network-wide decorrelation, correlations can be kept under control across all layers.

While our results are promising, we do recognize that most, though not quite all, of the performance gains of decorrelation via DBP in SAC is displayed when a smaller batch size is being used relative to the baseline. It should be explored further why decorrelation allows for smaller batch sizes and how robust this finding is in other contexts. Potentially, the reduction of redundant information could help the algorithm learn from fewer samples.

Further investigation is needed into the extension of our findings to other environments, architectures, and RL algorithms.
Results were demonstrated on a subset of most difficult games in the Atari suite where human-level performance is not reachable with DQN~\citep{mnih2015human}. Moving away from the most challenging task settings could potentially provide further performance benefits to DSAC. Whereas SAC is an ideal choice as a complex state-of-the-art deep RL algorithm, using simpler algorithms may ease the analysis on the impact of decorrelation directly, while also strengthening the generalizability of results. Furthermore, benefits of decorrelation may be more impactful for control tasks on continuous action space settings, as credit assignment is more challenging in larger action spaces, making effective representation learning more crucial.

When demonstrating the advantages of using decorrelation in deep RL through DBP,  different hyperparameter settings were used per game and per algorithm. 
Whereas not sticking to one configuration of hyperparameters may signal an inconsistency on the impact of decorrelation, the best hyperparameter configurations from the search for both the SAC baseline and DSAC were used to compare their performances in each game. The best conditions were thus ensured for not only DSAC but also the baseline to justify any improvements made over it, so that it is not merely explained away by a suboptimal choice of hyperparameters for the baseline algorithm. Moreover, because decorrelation is specific to data in our method, different games likely need different learning rates for the decorrelation process to be effective, as the hyperparameter search also demonstrated. However, the fact that decorrelation is specific to data could potentially have a negative impact on generalization.
One limitation could be the use of learning rates and batch sizes found as a result of the hyperparameter search conducted, which additionally included decorrelation in Q-networks for the DSAC algorithm, unlike the main results plotted because of the negligible impact of Q-networks' decorrelation for performance (see Appendix~\ref{sec.gridsearch}). While this may hint at suboptimality in our results, it suggests a possibility of finding even better hyperparameter configurations should the decorrelation in Q-networks already have been omitted in the search, implying a potential improvement on the results and conclusions drawn.

Results might be further improved  by going beyond mere decorrelation and applying whitening to the inputs, as whitening has been shown to benefit training and convergence of DNNs~\citep{luo2017learningdeeparchgwnn, huangi2018decorrbatchnorm, desjardins2015naturalnn}. Whitening constrains layer inputs to have unit variance, which can be achieved by the DBP algorithm by imposing the correlation matrix to be the identity matrix, and has been already shown to lead to improved performance over merely decorrelated inputs~\citep{dalm2024efficientdeeplearningdecorrelated}.

The positive impact of decorrelation on deep RL goes beyond training speed and reward benefits that signal effective representation learning and credit assignment. 
By decreasing the notoriously long training time for RL through addressing its issue of sample efficiency, decorrelation may reduce carbon emissions. 
As the network's decorrelation loss is essentially a measure of how familiar the states are, it might be used as a metric to guide exploration, improving sample efficieny further.
While a mechanism to promote exploration is already existent in SAC through the entropy maximization objective, other RL algorithms may benefit highly from such a exploration encouraging decorrelation signal.
Another possible application of decorrelation is explainable AI~\citep{rudin2019stopexplainingblackbox, ras2022explainabledl}, as decorrelated features remove confounding effects and make it easier to attribute a model's behaviour to a specific input feature.
The ability to decorrelate can additionally endow a network with predictive capabilities, as it informs us how surprising a newly sampled input is, which can be incorporated as a signal to help minimize sensory surprise, 
following after~\citet{küçükoğlu2024efficient} that models predictive processing~\citep{clark2013whatever} for effective control with RL, inspired by the theory of sensory information processing in the brain called predictive coding~\citep{srinivasan1982predictive,friston2005theory, huang2011}. In fact, neural processing has been shown to be engaged in input decorrelation~\citep{bell1997indepdendentcomponents, franke2017inhibition, pitkow2012decorrefficientcoding, segal2015decorrofretinalresponsebyfixationaleyemove}, thus further demonstrating the value of a decorrelation mechanism for developing biologically plausible efficient control algorithms.

Going beyond the domain of machine learning, decorrelation could be beneficial for intersecting disciplines like neurotechnology, specifically for the development of cortical visual neural prostheses. \citet{bengio2009slowdecorrelatedfeatures} proposed that decorrelation partly explains the functional behaviour of V1 complex cells. This hypothesis could be relevant for developing biologically inspired image processing approaches when modelling optimal encoding strategies that generate stimulation parameters for the visual cortex to evoke targeted perceptions in blind people. An adaptive task-based dynamic optimization approach to neuroprosthetic vision with end-to-end deep RL, for example as in~\citet{kucukoglu22tophos}, could therefore benefit from such a decorrelation mechanism. Furthermore, incorporation of eye movement information when developing image processing strategies for cortical neural prostheses has been found crucial in improving mobility outcomes for users ~\citep{de_Ruyter_van_Steveninck2024gazecontingent}. Decorrelation could help in modelling the incorporation of eye movement information for this purpose in a biologically plausible and efficient manner, building on the observation that fixational eye movements reduce correlations in retinal output~\citep{segal2015decorrofretinalresponsebyfixationaleyemove}.

Decorrelation targets a crucial issue in machine learning, which may negatively affect learning in the presence of highly correlated features. This is especially a prominent issue in deep RL, where training data is highly correlated due to the sequential nature of the learning framework. Therefore, future work can investigate a range of directions mentioned above to reap further benefits of decorrelation in deep RL, machine learning and beyond.

\section*{Acknowledgements}
This work has received funding from the European Union’s Horizon 2020 research and innovation programme under grant agreement No 899287 (project NeuraViPeR). 

\bibliographystyle{apalike}
\bibliography{main}

\begin{thebibliography}{}

\bibitem[Ahmad, 2024]{ahmad2024correlations}
Ahmad, N. (2024).
\newblock Correlations are ruining your gradient descent.
\newblock {\em ArXiv preprint}, ArXiv:2407.10780.

\bibitem[Ahmad et~al., 2023]{ahmad2023constrained}
Ahmad, N., Schrader, E., and van Gerven, M. (2023).
\newblock Constrained parameter inference as a principle for learning.
\newblock {\em Transactions on Machine Learning Research}.

\bibitem[Bell and Sejnowski, 1997]{bell1997indepdendentcomponents}
Bell, A.~J. and Sejnowski, T.~J. (1997).
\newblock The “independent components” of natural scenes are edge filters.
\newblock {\em Vision Research}, 37(23):3327--3338.

\bibitem[Bengio and Bergstra, 2009]{bengio2009slowdecorrelatedfeatures}
Bengio, Y. and Bergstra, J. (2009).
\newblock Slow, decorrelated features for pretraining complex cell-like networks.
\newblock In Bengio, Y., Schuurmans, D., Lafferty, J., Williams, C., and Culotta, A., editors, {\em Advances in Neural Information Processing Systems}, volume~22. Curran Associates, Inc.

\bibitem[Christodoulou, 2019]{christodoulou2019sacdiscrete}
Christodoulou, P. (2019).
\newblock Soft actor-critic for discrete action settings.
\newblock {\em ArXiv preprint}, ArXiv:1910.07207.

\bibitem[Clark, 2013]{clark2013whatever}
Clark, A. (2013).
\newblock {Whatever next? Predictive brains, situated agents, and the future of cognitive science}.
\newblock {\em Behavioral and Brain Sciences}, 36(3):181--204.

\bibitem[Dalm et~al., 2024]{dalm2024efficientdeeplearningdecorrelated}
Dalm, S., Offergeld, J., Ahmad, N., and van Gerven, M. (2024).
\newblock Efficient deep learning with decorrelated backpropagation.
\newblock {\em ArXiv preprint}, ArXiv:2405.02385.

\bibitem[de~Ruyter~van Steveninck et~al., 2024]{de_Ruyter_van_Steveninck2024gazecontingent}
de~Ruyter~van Steveninck, J., Nipshagen, M., van Gerven, M., Güçlü, U., Güçlütürk, Y., and van Wezel, R. (2024).
\newblock Gaze-contingent processing improves mobility, scene recognition and visual search in simulated head-steered prosthetic vision.
\newblock {\em Journal of Neural Engineering}, 21(2):026037.

\bibitem[Desjardins et~al., 2015]{desjardins2015naturalnn}
Desjardins, G., Simonyan, K., Pascanu, R., and Kavukcuoglu, K. (2015).
\newblock Natural neural networks.
\newblock In Cortes, C., Lawrence, N., Lee, D., Sugiyama, M., and Garnett, R., editors, {\em Advances in Neural Information Processing Systems}, volume~28. Curran Associates, Inc.

\bibitem[Franke et~al., 2017]{franke2017inhibition}
Franke, K., Berens, P., Schubert, T., Bethge, M., Euler, T., and Baden, T. (2017).
\newblock Inhibition decorrelates visual feature representations in the inner retina.
\newblock {\em Nature}, 542.

\bibitem[Friston, 2005]{friston2005theory}
Friston, K. (2005).
\newblock A theory of cortical responses.
\newblock {\em Philosophical Transactions of the Royal Society B: Biological sciences}, 360(1456):815--836.

\bibitem[Haarnoja et~al., 2018a]{haarnoja2018icml}
Haarnoja, T., Zhou, A., Abbeel, P., and Levine, S. (2018a).
\newblock Soft actor-critic: Off-policy maximum entropy deep reinforcement learning with a stochastic actor.
\newblock In Dy, J.~G. and Krause, A., editors, {\em ICML}, volume~80 of {\em Proceedings of Machine Learning Research}, pages 1856--1865. PMLR.

\bibitem[Haarnoja et~al., 2018b]{sac2018haarnoja}
Haarnoja, T., Zhou, A., Hartikainen, K., Tucker, G., Ha, S., Tan, J., Kumar, V., Zhu, H., Gupta, A., Abbeel, P., and Levine, S. (2018b).
\newblock Soft actor-critic algorithms and applications.
\newblock {\em ArXiv preprint}, ArXiv:1812.05905.

\bibitem[Huang et~al., 2023]{huang2023generalizabledecorsaliency}
Huang, S., Sun, Y., Hu, J., Guo, S., Chen, H., Chang, Y., Sun, L., and Yang, B. (2023).
\newblock Learning generalizable agents via saliency-guided features decorrelation.
\newblock In {\em Thirty-seventh Conference on Neural Information Processing Systems}.

\bibitem[Huang and Rao, 2011]{huang2011}
Huang, Y. and Rao, R. P.~N. (2011).
\newblock Predictive coding.
\newblock {\em WIREs Cognitive Science}, 2(5):580--593.

\bibitem[Huangi et~al., 2018]{huangi2018decorrbatchnorm}
Huangi, L., Yang, D., Lang, B., and Deng, J. (2018).
\newblock Decorrelated batch normalization.
\newblock In {\em Proceedings - 2018 IEEE/CVF Conference on Computer Vision and Pattern Recognition, CVPR 2018}, Proceedings of the IEEE Computer Society Conference on Computer Vision and Pattern Recognition, pages 791--800, United States. IEEE Computer Society.
\newblock Publisher Copyright: {\textcopyright} 2018 IEEE.; 31st Meeting of the IEEE/CVF Conference on Computer Vision and Pattern Recognition, CVPR 2018 ; Conference date: 18-06-2018 Through 22-06-2018.

\bibitem[Kaiser et~al., 2020]{kaiser2020simple}
Kaiser, L., Babaeizadeh, M., Milos, P., Osinski, B., Campbell, R.~H., Czechowski, K., Erhan, D., Finn, C., Kozakowski, P., Levine, S., Mohiuddin, A., Sepassi, R., Tucker, G., and Michalewski, H. (2020).
\newblock {Model based reinforcement learning for Atari}.
\newblock In {\em 8th International Conference on Learning Representations, {ICLR} 2020, Addis Ababa, Ethiopia, April 26-30, 2020}. OpenReview.net.

\bibitem[Küçükoğlu et~al., 2024]{küçükoğlu2024efficient}
Küçükoğlu, B., Borkent, W., Rueckauer, B., Ahmad, N., Güçlü, U., and van Gerven, M. (2024).
\newblock Efficient deep reinforcement learning with predictive processing proximal policy optimization.
\newblock {\em Neurons, Behavior, Data analysis, and Theory}, pages 1--24.

\bibitem[Küçükoğlu et~al., 2022]{kucukoglu22tophos}
Küçükoğlu, B., Rueckauer, B., Ahmad, N., de~Ruyter~van Steveninck, J., Güçlü, U., and van Gerven, M. (2022).
\newblock {Optimization of neuroprosthetic vision via end-to-end deep reinforcement learning}.
\newblock {\em International Journal of Neural Systems}, 32(11):2250052.

\bibitem[LeCun et~al., 2012]{lecun200efficientbp}
LeCun, Y.~A., Bottou, L., Orr, G.~B., and M{\"u}ller, K.-R. (2012).
\newblock Efficient backprop.
\newblock In Montavon, G., Orr, G.~B., and M{\"u}ller, K.-R., editors, {\em Neural Networks: Tricks of the Trade: Second Edition}, pages 9--48, Berlin, Heidelberg. Springer Berlin Heidelberg.

\bibitem[Lee et~al., 2023]{lee2023decorunsupreprlearn}
Lee, H., Lee, K., Hwang, D., Lee, H., Lee, B., and Choo, J. (2023).
\newblock On the importance of feature decorrelation for unsupervised representation learning in reinforcement learning.
\newblock In {\em Proceedings of the 40th International Conference on Machine Learning}, ICML'23. JMLR.org.

\bibitem[Linnainmaa, 1976]{linnainmaa1976bp}
Linnainmaa, S. (1976).
\newblock Taylor expansion of the accumulated rounding error.
\newblock {\em BIT}, 16(2):146–160.

\bibitem[Luo, 2017]{luo2017learningdeeparchgwnn}
Luo, P. (2017).
\newblock Learning deep architectures via generalized whitened neural networks.
\newblock In Precup, D. and Teh, Y.~W., editors, {\em Proceedings of the 34th International Conference on Machine Learning}, volume~70 of {\em Proceedings of Machine Learning Research}, pages 2238--2246. PMLR.

\bibitem[Machado et~al., 2018]{machado18}
Machado, M.~C., Bellemare, M.~G., Talvitie, E., Veness, J., Hausknecht, M., and Bowling, M. (2018).
\newblock {Revisiting the Arcade learning environment: Evaluation protocols and open problems for general agents}.
\newblock {\em Journal of Artificial Intelligence Research}, 61(1):523–562.

\bibitem[Mavrin et~al., 2019a]{mavrin2019effdecorgramian}
Mavrin, B., Graves, D., and Chan, A. (2019a).
\newblock {Efficient decorrelation of features using Gramian in reinforcement learning}.
\newblock {\em ArXiv preprint}, ArXiv:1911.08610.

\bibitem[Mavrin et~al., 2019b]{mavrin2019drlwithdecorr}
Mavrin, B., Yao, H., and Kong, L. (2019b).
\newblock Deep reinforcement learning with decorrelation.
\newblock {\em ArXiv preprint}, ArXiv:1903.07765.

\bibitem[Mnih et~al., 2013]{minh2013playingataridrl}
Mnih, V., Kavukcuoglu, K., Silver, D., Graves, A., Antonoglou, I., Wierstra, D., and Riedmiller, M.~A. (2013).
\newblock {Playing Atari with deep reinforcement learning}.
\newblock {\em ArXiv preprint}, ArXiv:1312.5602.

\bibitem[Mnih et~al., 2015]{mnih2015human}
Mnih, V., Kavukcuoglu, K., Silver, D., Rusu, A.~A., Veness, J., Bellemare, M.~G., Graves, A., Riedmiller, M., Fidjeland, A.~K., Ostrovski, G., et~al. (2015).
\newblock Human-level control through deep reinforcement learning.
\newblock {\em Nature}, 518(7540):529--533.

\bibitem[Pitkow and Meister, 2012]{pitkow2012decorrefficientcoding}
Pitkow, X. and Meister, M. (2012).
\newblock Decorrelation and efficient coding by retinal ganglion cells.
\newblock {\em Nature Neuroscience}, 15:628--35.

\bibitem[Ras et~al., 2022]{ras2022explainabledl}
Ras, G., Xie, N., van Gerven, M., and Doran, D. (2022).
\newblock Explainable deep learning: A field guide for the uninitiated.
\newblock {\em Journal of Artificial Intelligence Research}, 73.

\bibitem[Rudin, 2019]{rudin2019stopexplainingblackbox}
Rudin, C. (2019).
\newblock Stop explaining black box machine learning models for high stakes decisions and use interpretable models instead.
\newblock {\em Nature Machine Intelligence}, 1:206--215.

\bibitem[Schulman et~al., 2017]{schulman2017proximal}
Schulman, J., Wolski, F., Dhariwal, P., Radford, A., and Klimov, O. (2017).
\newblock Proximal policy optimization algorithms.
\newblock {\em ArXiv preprint}, ArXiv:1707.06347.

\bibitem[Segal et~al., 2015]{segal2015decorrofretinalresponsebyfixationaleyemove}
Segal, I.~Y., Giladi, C., Gedalin, M., Rucci, M., Ben-Tov, M., Kushinsky, Y., Mokeichev, A., and Segev, R. (2015).
\newblock Decorrelation of retinal response to natural scenes by fixational eye movements.
\newblock {\em Proceedings of the National Academy of Sciences}, 112(10):3110--3115.

\bibitem[Srinivasan et~al., 1982]{srinivasan1982predictive}
Srinivasan, M.~V., Laughlin, S.~B., and Dubs, A. (1982).
\newblock Predictive coding: A fresh view of inhibition in the retina.
\newblock {\em Proceedings of the Royal Society of London. Series B. Biological Sciences}, 216(1205):427--459.

\bibitem[Sutton and Barto, 2018]{Sutton1998rlintro}
Sutton, R.~S. and Barto, A.~G. (2018).
\newblock {\em Reinforcement Learning: An Introduction}.
\newblock The MIT Press, second edition.

\bibitem[Toromanoff et~al., 2019]{toromanoff2019deep}
Toromanoff, M., Wirbel, E., and Moutarde, F. (2019).
\newblock {Is deep reinforcement learning really superhuman on Atari? Leveling the playing field}.
\newblock {\em ArXiv preprint}, ArXiv:1908.04683.

\bibitem[Werbos, 1974]{werbos1974bp}
Werbos, P. (1974).
\newblock {\em Beyond Regression: New Tools for Prediction and Analysis in the Behavioral Sciences}.
\newblock PhD thesis, Harvard University, Cambridge, MA, USA.

\end{thebibliography}


\begin{appendices}

\section{Grid search for hyperparameters} 
The grid search for hyperparameters were based on DSAC where all trained networks are decorrelated, including the policy network and both Q-networks. The decorrelation learning rate for the two soft Q-networks were fixed at $1 \times 10^{-13}$, after a preliminary search showing it as a reasonable value. For the grid search, the BP-based SAC baseline runs were conducted based on similar architectural conditions to ensure equality in model sizes.

While the grid search results also involve decorrelation on the two soft Q-networks, for our main results the decorrelation procedure was not applied in the Q-networks, because of its negligible impact on overall performance compared to the decorrelation in the policy network, and also the increase in computational cost and training time with the decorrelation of two additional networks. 
Also for the main results plotted, the SAC baseline lacked the decorrelation matrices fixed as identity matrices that ensured equal model sizes in the hyperparameter search, in order to do justice to the SAC baseline in the main results by evaluating it without the additional time cost of a non-trained decorrelation process.

\label{sec.gridsearch}
\begin{figure*}[!ht]
\centering
\subfigure{\includegraphics[width=5cm]{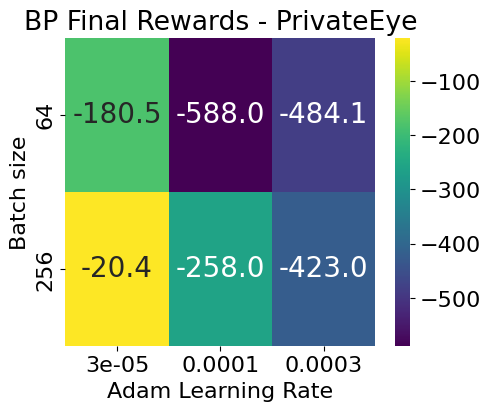}}
\subfigure{\includegraphics[width=4.8cm]{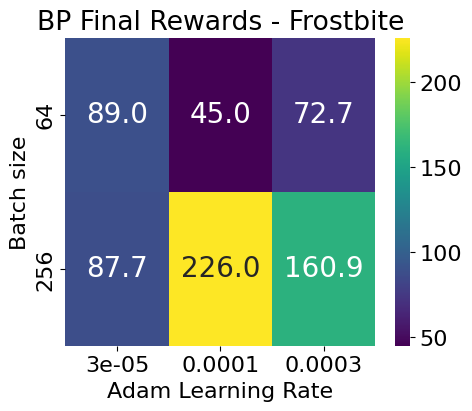}}
\subfigure{\includegraphics[width=4.8cm]{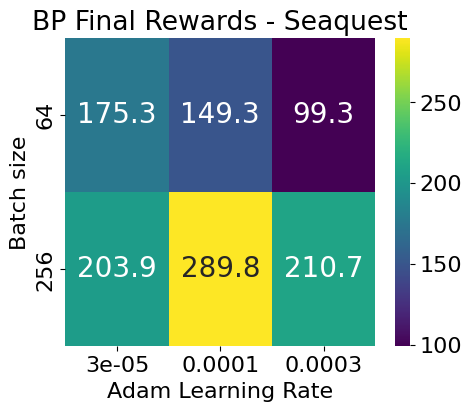}}
\subfigure{\includegraphics[width=5cm]{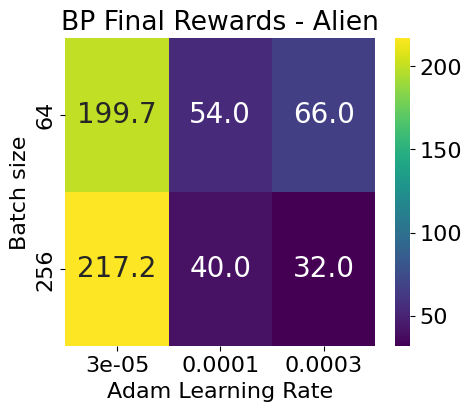}}
\subfigure{\includegraphics[width=5cm]{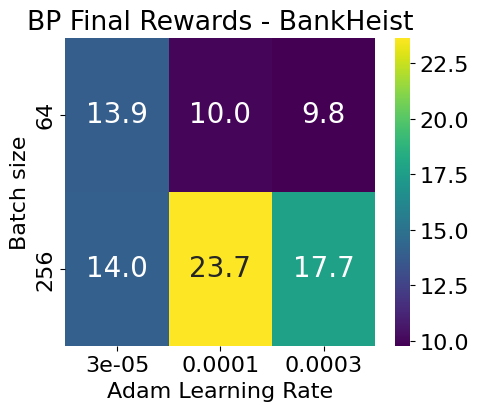}}
\subfigure{\includegraphics[width=5.4cm]{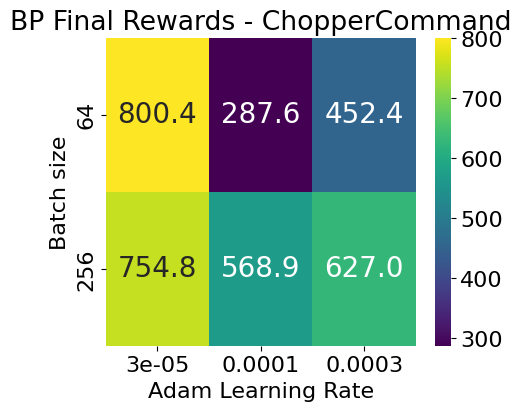}}
\subfigure{\includegraphics[width=5cm]{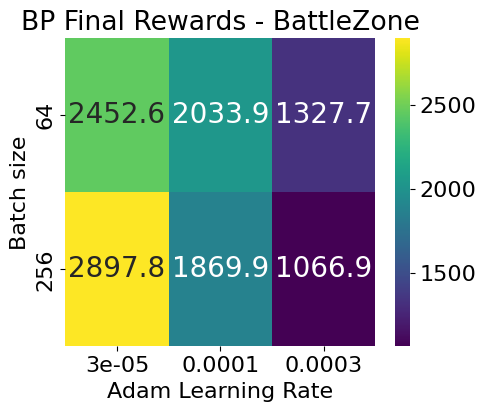}}
\caption{Search for the best hyperparameter configuration for the BP-based discrete SAC baseline. Values indicate the final reward average across five runs.}
\label{fig:bpgrid}
\end{figure*}

\newpage
\begin{figure*}[!ht]
\centering
\subfigure{\includegraphics[width=5.01cm]{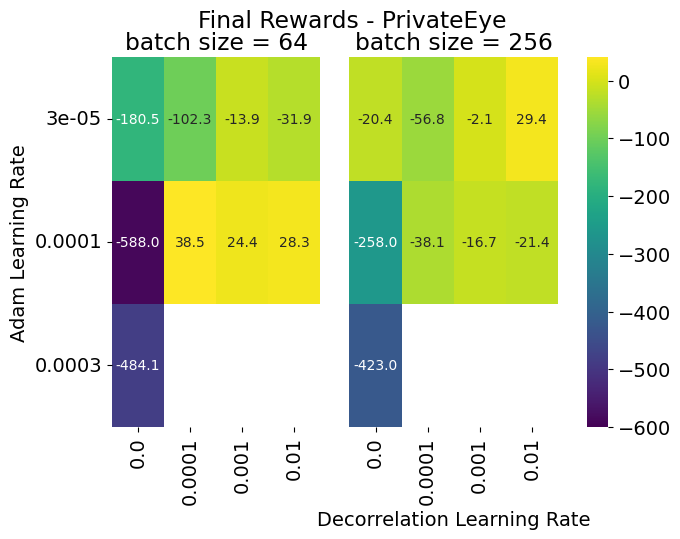}}
\subfigure{\includegraphics[width=5.01cm]{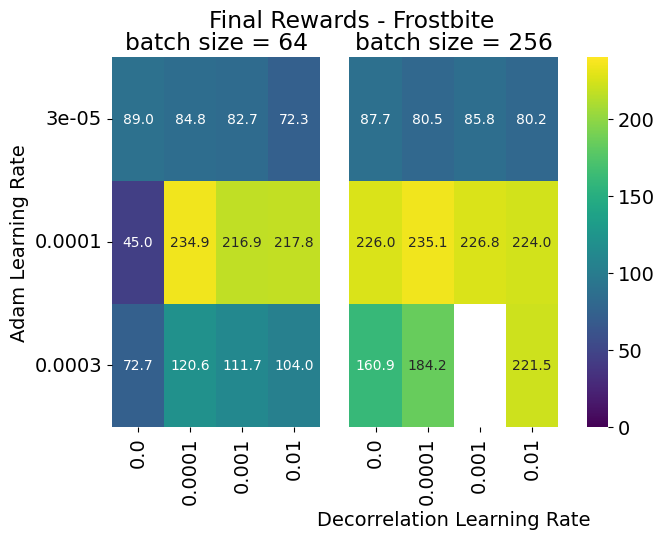}}
\subfigure{\includegraphics[width=5.01cm]{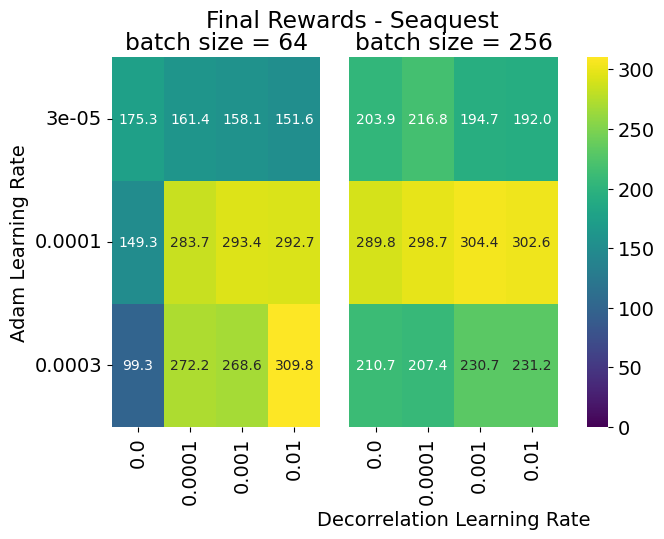}}
\subfigure{\includegraphics[width=5.15cm]{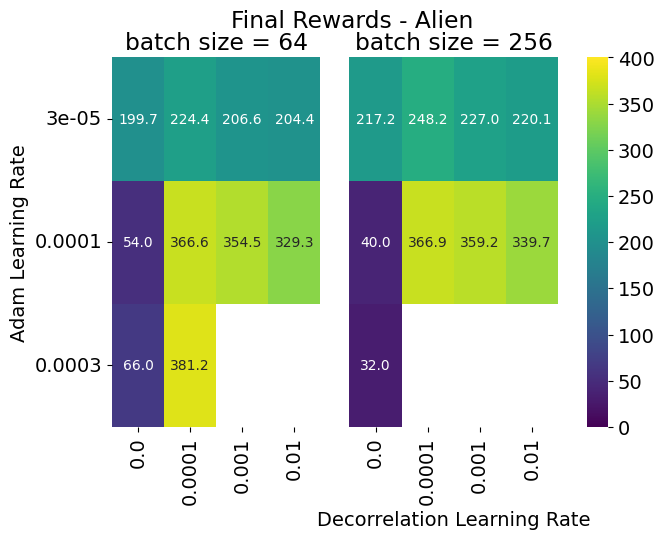}}
\subfigure{\includegraphics[width=5.15cm]{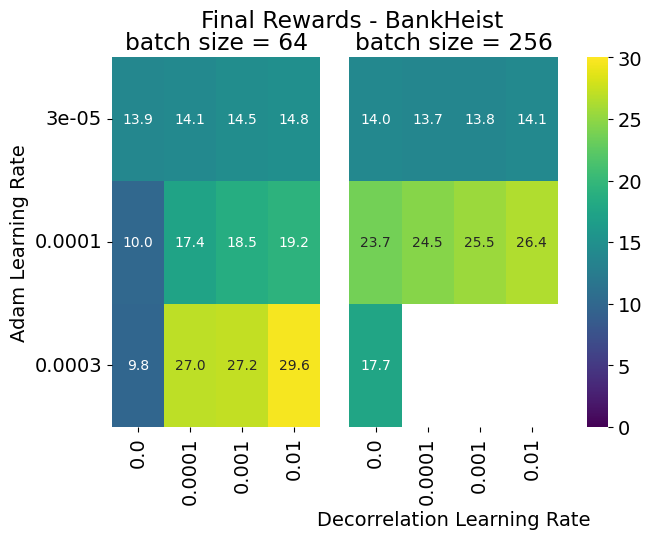}}
\subfigure{\includegraphics[width=5.15cm]{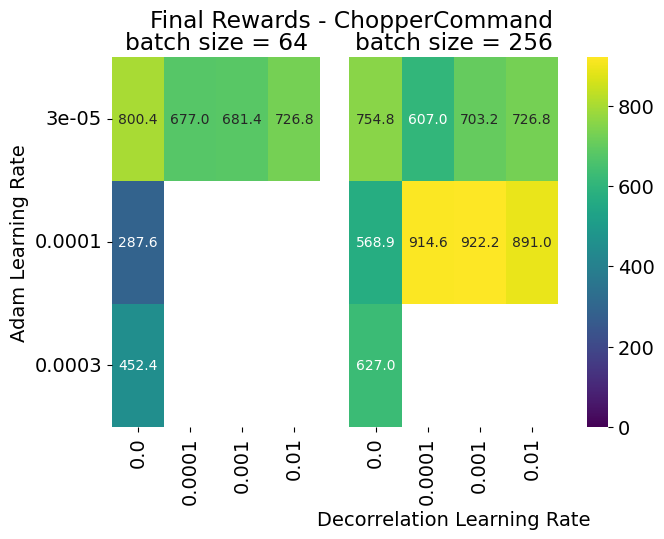}}
\subfigure{\includegraphics[width=5.15cm]{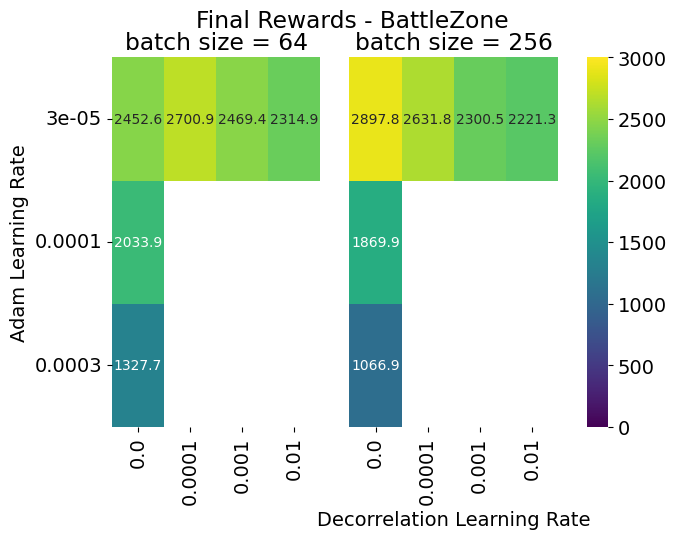}}
\caption{Search for the best hyperparameter configuration for DSAC. Values indicate the final reward average across five runs. A decorrelation learning rate of 0.0 indicate no decorrelation for any of the SAC networks, hence the BP-based SAC baseline. A decorrelation learning rate more than 0.0 indicate decorrelation also in the soft-Q-networks, yet with a fixed learning rate of $1 \times 10^{-13}$, hence the DBP-based DSAC algorithm. Empty fields indicate NaNs.}
\label{fig:dbpgrid}
\end{figure*}

\newpage
\newpage
\section{Hyperparameters for DSAC} \label{sec.hyperparams}
\begin{table}[!ht]
\label{tab:hyper}
\caption{Hyperparameters used in the DSAC agents. Majority of the SAC-related hyperparameters follow after~\citep{christodoulou2019sacdiscrete}, e.g. learning rate, optimizer, discount factor, GAE parameter, PPO clip range, or after~\citep{sac2018haarnoja}, e.g. epochs per batch, actor loss coefficient, critic loss coefficient.} 

\centering
\begin{small}
\begin{tabular}{l|l} 
Hyperparameter & Value \\
\hline 
Layers (both for actor and critic) & 3 convolutional + 2 fully-connected layers \\ 
Convolutional channels per layer & [32, 64, 64]\\
Convolutional kernel sizes per layer & [8, 4, 3] \\
Convolutional strides per layer & [4, 2, 1] \\
Convolutional padding per layer & [0, 0, 0] \\
Fully connected layer hidden units & [512, number of actions in the game] \\ 
Nonlinearity & Leaky ReLU \\
Scaling coefficient for convolutional DBP layers ($b$) & 9 \\

Optimizer & Adam \\ 
Discount factor $(\gamma)$ & 0.99 \\
Replay buffer size & 100000 \\
Initial random steps & 20000 \\
Entropy target & -dim(A)\\ 
Target smoothing coefficient ($\tau$) & 0.005\\ 
Target update interval & 1\\ 
Gradient steps & 1 \\ 
Frame stacking & 4 \\
Frame skipping & 4 \\
\end{tabular}
\end{small}
\end{table}

\section{Hardware and implementation details} \label{sec.hardware}

We programmed our implementation in Python using the PyTorch framework, and ran on the compute cluster Snellius using Nvidia A100 GPUs, provided by SURF.

\newpage
\section{Top configurations for hyperparameter search} \label{sec.topconfigurations}
\begin{table}[!ht]
\label{tab:bestconfigurations}
\caption{Reward and wall-clock time performance of the plotted algorithms, based on the best configuration of learning rates (lr) and batch sizes (bs) found per game as a result of the hyperparameter search for algorithms of BP-based SAC baseline and DBP-based DSAC. Decorrelation learning rate here refers to the one used for the policy network. 
Decorrelation learning rates for the soft-Q-networks were zero in the plotted algorithms, while the hyperparameter search was based on the case where it was kept fixed at $1 \times 10^{-13}$ for DSAC, which had negligible impact on results. 
In parentheses, the change for the respective performance metric compared to the SAC baseline is provided, with significant changes in reward levels being starred. Note that there may be further time or performance benefits if the algorithm was to be run on different hyperparameter configurations, since the hyperparameters were based on the search that included decorrelation in Q-networks as well, as opposed to the plotted results that lacked it.}

\centering
\begin{small}
\begin{tabular}{l|l|l|l|l|l} 
Environment-Algorithm & Lr & Bs & Decorr. lr ($\eta$)& Final score & Wall-clock time (min.)\\
\hline 
\hline  
PrivateEye-SAC & $3 \times 10^{-5}$ & 256  & - &  -51.7 & 239 \\ 
PrivateEye-DSAC & $1 \times 10^{-4}$ & 64 & $1 \times 10^{-4}$ &  24.1 (+147\%) & 133 (-44\%) \\ 
\hline 
Frostbite-SAC & $1 \times 10^{-4}$ & 256 & - & 225.6 & 129\\ 
Frostbite-DSAC & $1 \times 10^{-4}$ & 256 & $1 \times 10^{-4}$ & 226.7 (+0\%) & 139 (+8\%) \\
\hline 
Seaquest-SAC & $1 \times 10^{-4}$ & 256  & - & 282.0 & 140\\ 
Seaquest-DSAC & $3 \times 10^{-4}$ & 64 & 0.01 & 299.6 (+6\%*) & 79 (-44\%) \\ 
\hline 
Alien-SAC & $3 \times 10^{-5}$ & 256  & - &  191.2 & 137\\ 
Alien-DSAC & $3 \times 10^{-4}$ & 64 & $1 \times 10^{-4}$ & 356.3 (+86\%*) & 59 (-57\%) \\ 
\hline 
BankHeist-SAC & $1 \times 10^{-4}$ & 256  & - & 22.5 & 104\\ 
BankHeist-DSAC & $3 \times 10^{-4}$ & 64 & $1 \times 10^{-2}$ & 28.1 (+25\%) & 63 (-39\%) \\ 
\hline 
ChopperCommand-SAC & $3 \times 10^{-5}$ & 64  & - & 818.8  &66\\
ChopperCommand-DSAC & $1 \times 10^{-4}$ & 256 & $1 \times 10^{-3}$ &  912.6 (+11\%) & 256 (+288\%) \\ 
\hline 
BattleZone-SAC & $3 \times 10^{-5}$ & 256  &  - & 2216.1 & 246 \\ 
BattleZone-DSAC & $3 \times 10^{-5}$ & 64 & $1 \times 10^{-4}$  & 2504.7 (+13\%) & 58 (-76\%) \\ 
\end{tabular}
\end{small}
\end{table}

\newpage
\section{Rewards for DSAC with policy network decorrelation} 
\label{sec.mainresults}
\begin{figure*}[!ht]
\centering
\includegraphics[width=\textwidth]{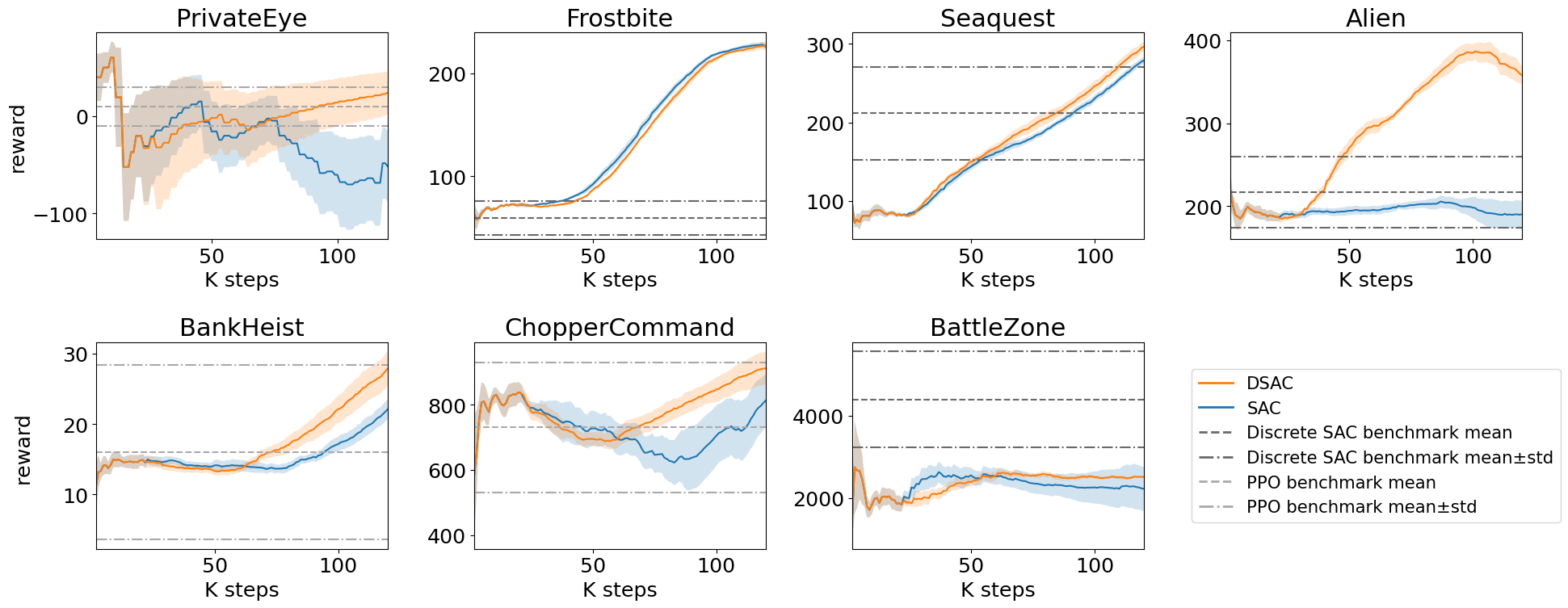}
\caption{Reward curves per game for top configurations of SAC baseline vs. DSAC with decorrelation in the policy network only. Discrete SAC benchmark scores are based on~\citet{christodoulou2019sacdiscrete}, and are reported where available for a comparison. For the remaining games, scores of PPO for the Atari 100k benchmark, as reported by~\citep{kaiser2020simple} are provided, as PPO is another state-of-the-art actor-critic algorithm, albeit learning on-policy~\citep{schulman2017proximal}. Comparison to algorithms in the literature demonstrates that the results of our BP-based SAC baseline are aligned with the previously reported BP-based benchmarks for most of the games. Differences may be due to use of different seeds.} 
\label{fig:rewardadv}
\end{figure*}

\newpage
\section{SAC losses for DSAC with policy network decorrelation} \label{sec.losscurves}

\begin{figure*}[!htb]
\centering
\includegraphics[width=\textwidth]{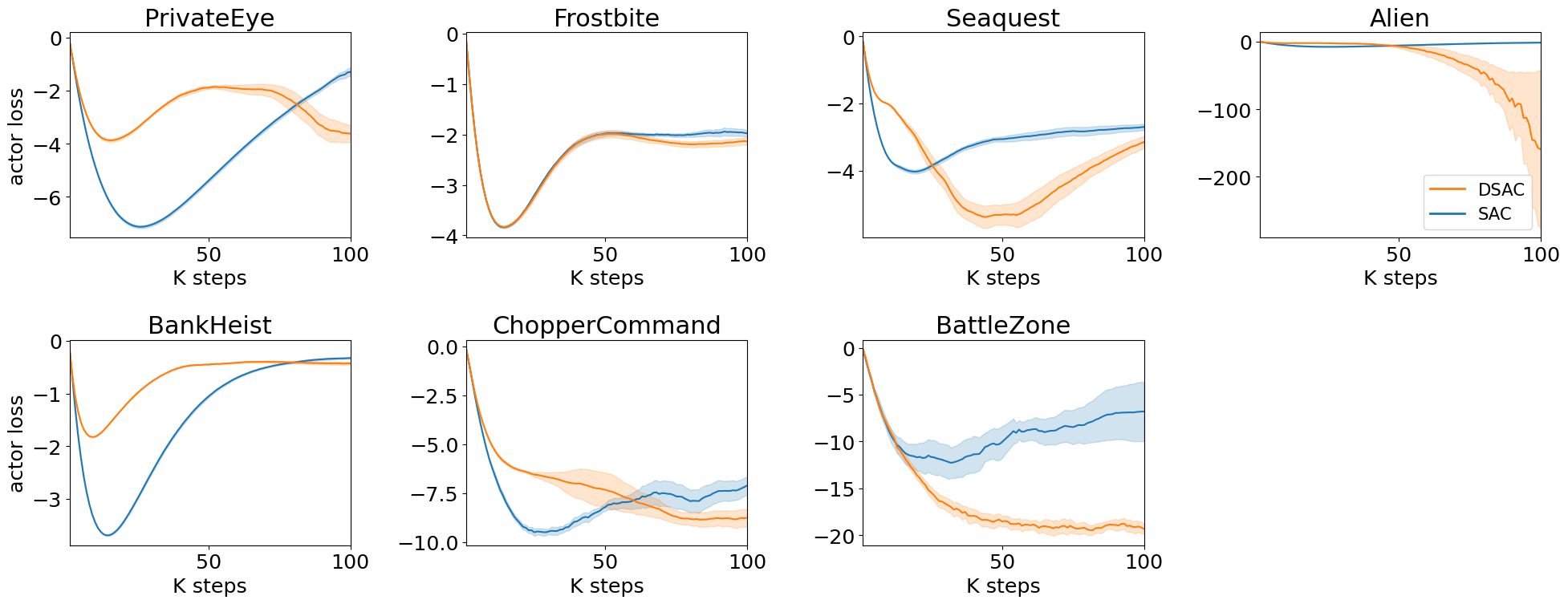}
\caption{Decorrelation helps actor loss or policy network’s loss to stay around minimal values
especially at the beginning of the training, where SAC baseline tends to have higher values
more quickly.}
\label{fig:actorloss}
\end{figure*}

\begin{figure*}[!htb]
\centering
\includegraphics[width=\textwidth]{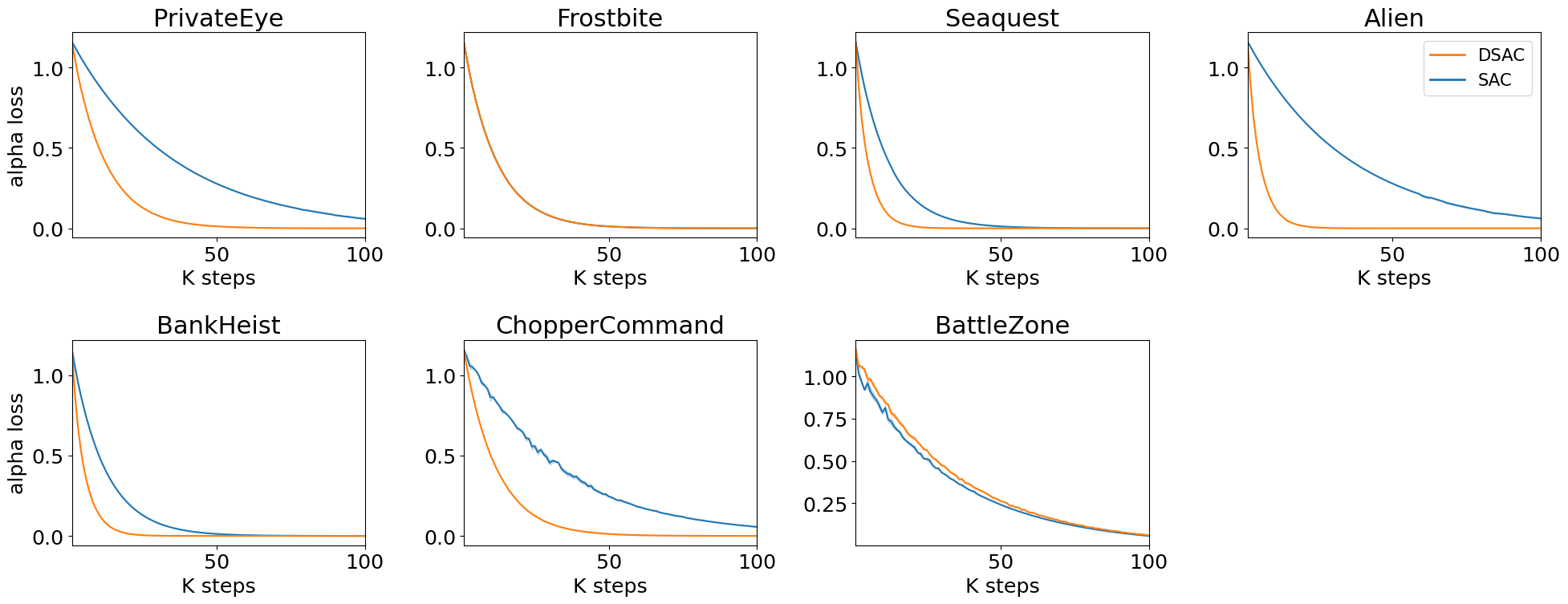}
\caption{Decorrelation enables a faster convergence for the minimization of alpha loss that serves the entropy maximization objective encouraging exploration and stability.}
\label{fig:alphaloss}
\end{figure*}

\begin{figure*}[!htb]
\centering
\includegraphics[width=\textwidth]{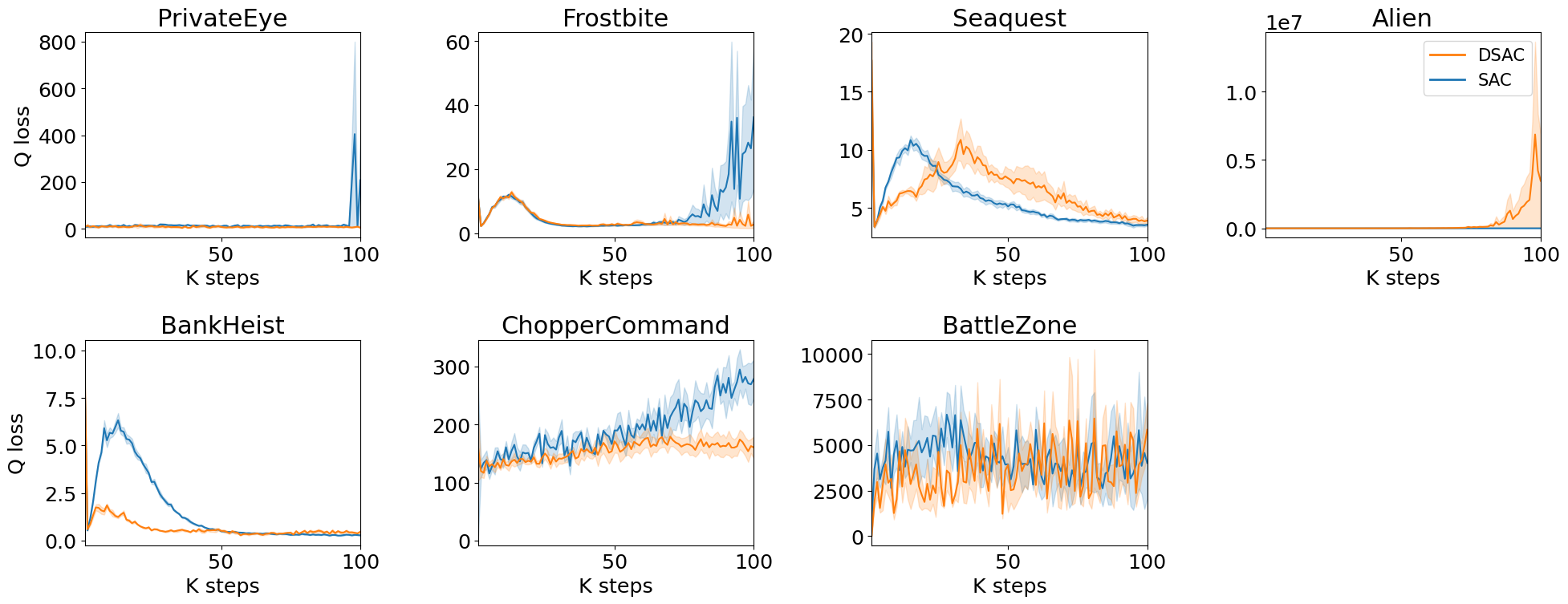}
\caption{Average Q-loss from the two soft Q-networks.}
\label{fig:qloss}
\end{figure*}

\newpage

\section{DSAC with decorrelation in all trained networks} 
\label{sec.resultsalldecorrelated}
Here we show the performance comparison of top performers of the BP-based SAC baseline and DBP-based DSAC algorithms based on the hyperparameter search where decorrelation is applied across all trained networks, including the policy network and the two Q-networks. Because of the decorrelation process being applied in three networks, hence increased computation, time advantage of the DBP-based DSAC is less pronounced than in the case of only the policy network being decorrelated, which the main results demonstrated. 

\begin{figure*}[!ht]
\centering
\includegraphics[width=\textwidth]{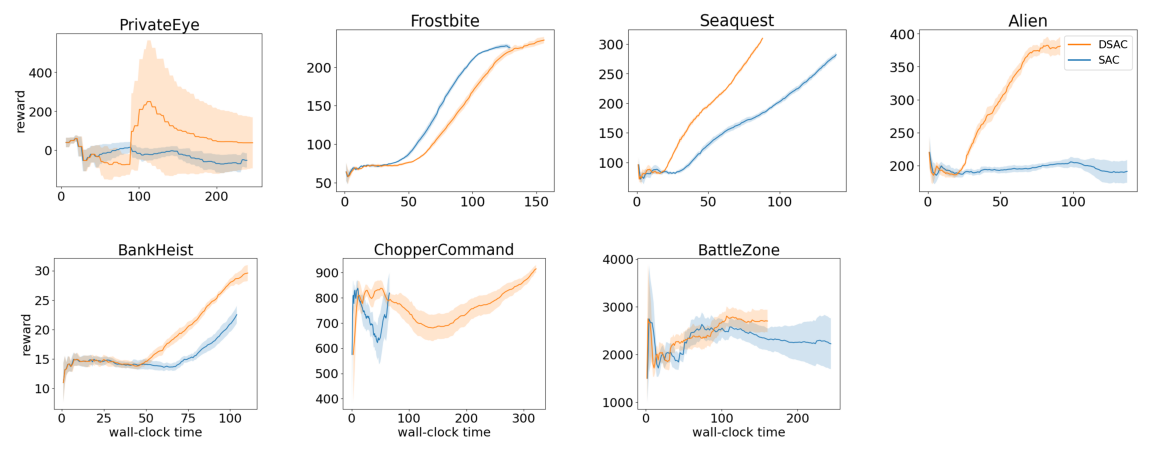}
\caption{Time advantage of the DBP-based DSAC algorithm (with decorrelation in all trained networks) in speeding up the training in 
3
of the 7 games tested for up to 
37\%
, while performance is maintained in all 7, with even a significant improvement via the DSAC in 3 games (Seaquest, Alien, BankHeist). Both algorithms are demonstrated with their best hyperparameter configuration for each game, with results averaged over 5 seeds and shaded area showing standard error of the mean. DSAC capitalizes on lower values for the batch size (64) hyperparameter against the SAC baseline in all the games it is faster on.
For ChopperCommand, algorithm is slower due to higher batch size of the best configurations of DSAC (256) against SAC baseline (64) for this game. Only for Frostbite, the batch size of DSAC is identical to SAC's (256).} 
\label{fig:timeadvantage_alldecor}
\end{figure*}

\begin{figure*}[!ht]
\centering
\includegraphics[width=\textwidth]{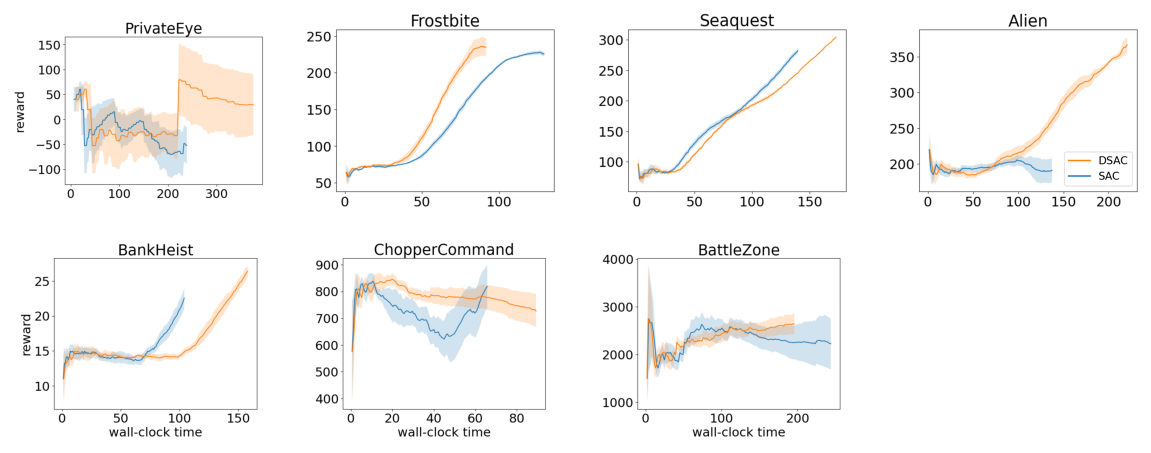}
\caption{Wall-clock time and reward performance of the highest ranking configurations in the other batch size than the top performer of DSAC (with all trained networks decorrelated) per each algorithm per game. Note that the batch sizes for the DSAC and SAC baseline per game are identical here for all games, except Frostbite where the DSAC algorithm has a lower batch size (64). The identical batch sizes for the majority of these games are 256, whereas 64 for ChopperCommand. Lower batch size for DSAC provides additional reduction in training time for Frostbite. 
At an identical batch size to the SAC baseline, both the training time advantage and the reward performance level is maintained for BattleZone, despite a slight reduction in the level of time advantage. 
On the other hand, while DSAC-based Seaquest, Alien, PrivateEye and BankHeist lose their training time advantage at an identical batch size, and BankHeist also its performance advantage, Seaquest and Alien still outperform the SAC baseline, whereas PrivateEye similarly maintains its performance levels.}
\label{fig:timeadv_2ndbs}
\end{figure*}

\end{appendices}

\end{document}